\documentclass[letterpaper, 10 pt, journal, twoside]{IEEEtran}
\usepackage{amsmath,amsfonts}
\usepackage{algorithmic}
\usepackage{algorithm}
\usepackage{array}
\usepackage[caption=false,font=normalsize,labelfont=sf,textfont=sf]{subfig}
\usepackage{textcomp}
\usepackage{stfloats}
\usepackage{url}
\usepackage{verbatim}
\usepackage{graphicx}
\usepackage{cite}
\usepackage{multicol} % assumes amsmath package installed
\usepackage{multirow} % assumes amsmath package installed
\usepackage{relsize}
\usepackage{hhline}
\usepackage{float}
\usepackage[table,xcdraw]{xcolor}
\usepackage{diagbox}
\usepackage{hyperref}

\newcommand{\twodots}{\mathinner {\,\ldotp \ldotp\,}}
\newcommand{\argmax}{\mathop{\mathrm{argmax}}} 

\begin{document}
\bstctlcite{IEEEexample:BSTcontrol}

\title{Visual Spatial Attention and Proprioceptive Data-Driven Reinforcement Learning for Robust Peg-in-Hole Task Under Variable Conditions}

\author{Andr\'e Yuji Yasutomi$^{1,2}$, Hideyuki Ichiwara$^{1,2}$, Hiroshi Ito$^{1,2}$, Hiroki Mori$^{3}$ and Tetsuya Ogata$^{2,4}$% <-this % stops a space
\thanks{Manuscript received: October 1, 2022; Revised December 13, 2022; Accepted January 29, 2023.} %Use only for final RAL version
\thanks{This paper was recommended for publication by Editor Hyungpil Moon upon evaluation of the Associate Editor and Reviewers' comments.
This work was supported by Hitachi, Ltd.} %Use only for final RAL version
\thanks{$^{1}$Andr\'e Yuji Yasutomi, Hideyuki Ichiwara and Hiroshi Ito are with the R\&D Group, Hitachi, Ltd, Japan 
{\tt\footnotesize andre.yasutomi.ss@hitachi.com}} %
\thanks{$^{2}$Andr\'e Yuji Yasutomi, Hideyuki Ichiwara, Hiroshi Ito and Tetsuya Ogata are with the Graduate School of Fundamental Science and Engineering, Waseda University, Japan
{\tt\footnotesize ogata@waseda.jp}}%
\thanks{$^{3}$Hiroki Mori is with the Future Robotics Organization, Waseda University, Japan
{\tt\footnotesize  mori@idr.ias.sci.waseda.ac.jp}}%
\thanks{$^{4}$Tetsuya Ogata is with the Waseda Research Institute for Science and Engineering (WISE), Waseda University, Japan}%
\thanks{Digital Object Identifier (DOI): \href{https://doi.org/10.1109/LRA.2023.3243526}{10.1109/LRA.2023.3243526}}
}

\markboth{IEEE Robotics and Automation Letters. Preprint Version. Accepted January, 2023}%
{Yasutomi \MakeLowercase{\textit{et al.}}: Visual Spatial Attention and Proprioceptive Data-Driven RL for Robust Peg-in-Hole Task}

\IEEEpubid{
\begin{tabular}[t]{@{}c@{}}~\copyright~2023 IEEE. Personal use of this material is permitted. Permission from IEEE must be obtained for all other uses, \\
                                    in any current or future media, including reprinting/republishing this material for advertising or promotional purposes, creating new collective works, \\ for resale or redistribution to servers or lists, or reuse of any copyrighted component of this work in other work\end{tabular}}

\maketitle

\begin{abstract}
  Anchor-bolt insertion is a peg-in-hole task performed in the construction field for holes in concrete. Efforts have been made to automate this task, but the variable lighting and hole surface conditions, as well as the requirements for short setup and task execution time make the automation challenging. In this study, we introduce a vision and proprioceptive data-driven robot control model for this task that is robust to challenging lighting and hole surface conditions. This model consists of a spatial attention point network (SAP) and a deep reinforcement learning (DRL) policy that are trained jointly end-to-end to control the robot. The model is trained in an offline manner, with a sample-efficient framework designed to reduce training time and minimize the reality gap when transferring the model to the physical world. Through evaluations with an industrial robot performing the task in 12 unknown holes, starting from 16 different initial positions, and under three different lighting conditions (two with misleading shadows), we demonstrate that SAP can generate relevant attention points of the image even in challenging lighting conditions. We also show that the proposed model enables task execution with higher success rate and shorter task completion time than various baselines. Due to the proposed model's high effectiveness even in severe lighting, initial positions, and hole conditions, and the offline training framework's high sample-efficiency and short training time, this approach can be easily applied to construction.
\end{abstract}

\begin{IEEEkeywords}
Robotics and automation in construction; reinforcement learning; deep learning for visual perception.
\end{IEEEkeywords}

\section{Introduction}
\IEEEPARstart{A}{nchor-bolt} insertion is a peg-in-hole task for holes in concrete that is extensively performed in the construction field \cite{anchor}. Since it is repetitive, dirty, and dangerous \cite{andre_icra2021}, efforts have been made to automate this task with industrial robots \cite{andre_icra2021,andre_sii2022}. However, automating anchor-bolt insertion is particularly difficult in this field because the robot must be able to adapt to variable conditions such as drastic changes in lighting caused by natural light or various light sources, and different hole surface conditions due to the brittle nature of concrete. Furthermore, the robot should be setup (e.g., trained) and be able to execute the task in short time in order to reach a lead time close to the time taken by humans.

To this end, in our prior study \cite{andre_icra2021}, we proposed a peg-in-hole strategy that used a deep neural network (DNN) trained via deep reinforcement learning (DRL) to generate discrete robot motions that adapted to different hole conditions. This method used force, torque, and robot displacement as input to avoid visual-detection problems caused by harsh lighting conditions. While the method was effective for finding holes, it required long time to be executed.

\IEEEpubidadjcol

The use of visual feedback with image feature detection algorithms is a promising approach to reduce the task completion time because it enables direct estimation of the hole position to effectively guide the robot. However, traditional algorithms such as template matching \cite{template} struggle to perform well in poorly lit conditions. Subsequent approaches \cite{im1,deep_learning,yolov4} involving deep convolutional neural networks (CNNs) have been used successfully in challenging lighting conditions, but they require large amounts of data to learn to be invariant to irrelevant image features \cite{deep_learning}.

Another strategy to reduce task completion time is to extensively train DRL algorithms in simulation environments and transfer the trained model to the real world \cite{sim2real_survey}. Some peg-in-hole approaches \cite{pih_beltran_rl,peginhole_gazebosim1} have accomplished this transfer by reducing the reality gap using techniques such as domain randomization \cite{domain_randomization}. However, in the present study, it is difficult to model the friction of concrete to be close enough to reality to use these approaches successfully. This is because the concrete's brittleness makes the wall surface to be inconsistent, with areas of high friction coefficient that prevent the peg from sliding, or make it slide unsteadily that results in excessively noisy forces and torques.

\begin{figure*}[t!]
  \begin{center}
  \includegraphics[width=\linewidth]{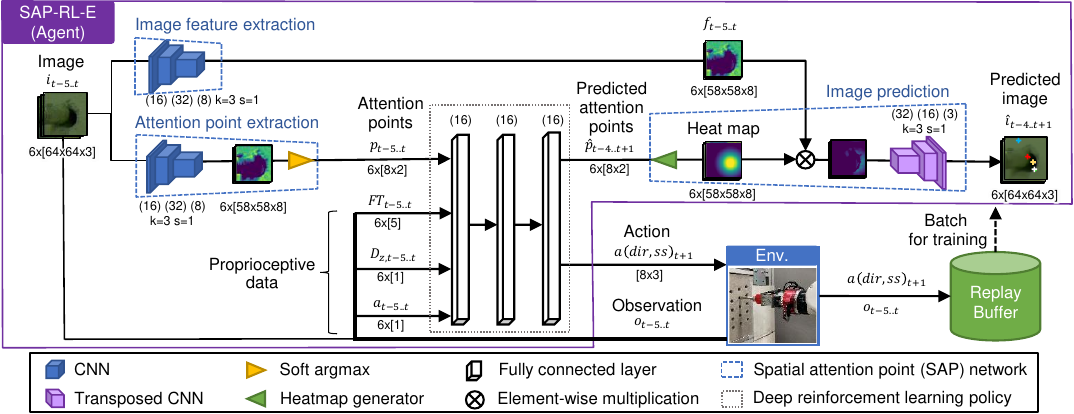}
  \caption{Proposed model structure and training architecture. For the CNNs, values in parentheses are the channel sizes, \textit{k} is kernel size, and \textit{s} is stride. For the fully connected layers, values in parentheses are the number of nodes. Values in brackets are the data sizes. $FT$ are the forces and torques, $D_z$ is robot displacement toward the wall, and $a(dir,ss)$ is the discrete robot action which is dependent on the direction $dir$ and step size $ss$.}
  \label{fig:saprle2e}
  \end{center}
\end{figure*}

Recent deep learning approaches integrate spatial attention, autoencoders, and robot motion generation networks \cite{sacae,dsae,ichiwara_arxiv,ichiwara_icra22} to learn to generate motion based on image features as well as task-dependent image features. These approaches have proven to be sample-efficient and generalize well to different image conditions. Using these approaches as our basis, we propose a sample-efficient vision-based robot control model for performing peg-in-hole tasks in concrete holes (Fig. \ref{fig:saprle2e}). The model is robust to challenging lighting conditions (with misleading shadows), is trained in an offline manner and in a short time, and contributes to the improvement of the task success rate and completion time.

Our main contributions are as follows: (1) modifications of a spatial attention point network (SAP), introduced in prior work \cite{ichiwara_arxiv} to extract important points (referred to as ``attention points'') of images even without a recurrent layer; (2) the introduction of a model that consists of SAP integrated to a DRL policy, and a method to jointly train them end-to-end to teach SAP to generate task-specific attention points, and the DRL policy for generating robot motion based on the attention points; (3) an evaluation of SAP and the proposed model for images with misleading shadows; and (4) an offline framework to train models in short time, with minimal data extraction from the environment and a minimal reality gap when transferring the models to the actual field.

\subsection{Related work} 

\subsubsection{Peg-in-hole with DRL}

DRL has been commonly applied to real robots to perform peg-in-hole tasks without visual input \cite{pih_inoue_rl,pih_beltran_rl}. These ``blind'' approaches typically use force, torque, and position feedback to train DRL policies (e.g., deep q-network (DQN) \cite{pih_inoue_rl}, deep deterministic policy gradients (DDPG) \cite{peginhole_vision_dl2}, and soft actor critic (SAC) \cite{pih_beltran_rl}) to control the robot by outputting position/orientation subgoals, force/moment subgoals, or parameters for robot compliance control \cite{pih_beltran_rl}. Although these approaches are effective, they rely on simulators and sliding the peg around the hole to search for it, which is difficult for holes in concrete due to the concrete's brittleness and high friction coefficient.

DRL has also been applied to peg-in-hole tasks with visual feedback \cite{peginhole_vision_dl1,peginhole_vision_dl2,spatialattention_2016}. However, these studies rely on cameras fixed apart from the robot end effector and require prior motion information from classical controllers, humans demonstrations \cite{peginhole_vision_dl2,spatialattention_2016}, or extensive simulation \cite{peginhole_vision_dl1} to train safe and sample-efficient visuomotor policies. In addition, they did not fully evaluate the robustness of their approaches in challenging lighting variations.

In contrast, in this study we move the robot in a hopping manner to avoid the friction against concrete, and we avoid simulators by extracting small amounts of proprioceptive data and images (from a camera fixed to the end effector) from the field for offline training. We show that a model trained with this motion and training method can enable a robot to accomplish a peg-in-hole task in harsh lighting conditions.

\subsubsection{Attention for motion generation} 
Several models have been proposed that use visual attention to extract image features to provide highly relevant input for motion controllers \cite{spatialattention_2016, dsae,ichiwara_arxiv, ichiwara_icra22, spatial3d}. Levine et al. \cite{spatialattention_2016, dsae} proposed the use of CNNs and soft argmax to obtain position coordinates in an image for image feature extraction. They showed that jointly training the image feature extraction model and a reinforcement learning (RL) policy end-to-end results in higher success rates than training them separately. Their method enabled a robot to accomplish multiple tasks (e.g., coat hanging and cube fitting) but it required at least 15000 images, 3 to 4 hours of training time, and does not generalize to drastically different settings, especially when visual distractors occlude the target objects.

Ichiwara et al. \cite{ichiwara_arxiv, ichiwara_icra22} proposed a method of jointly training a point-based attention mechanism that predicts attention points based on image reconstruction and a long-short term memory (LSTM) layer end-to-end to generate motion. Their method enabled a robot to perform simple pick and place tasks and a zipper closing task with variable backgrounds, light brightness, and distractor objects. Their method requires small amounts of data, but it also requires expert human demonstrations (for imitation learning) and high computation cost due to the LSTM layer (up to 3 h of training in 2 GPUs of 16GB of memory). Moreover, it was not tested with misleading shadows.

In this study, we propose a visual attention-based model which has low computational cost and is trained with small amounts of data acquired automatically from the environment (no expert demonstration). By evaluating this model in an experimental setup that replicates a construction site, which includes misleading shadows and variable target surface conditions, we demonstrate that it is able to effectively conduct a peg-in-hole task even in these conditions.

\section{Methods}\label{sec:methods}
\subsection{Hole search and peg insertion strategy}\label{subsec:holesearch}

\begin{figure}[t]
  \begin{center}
  \includegraphics[width=\columnwidth]{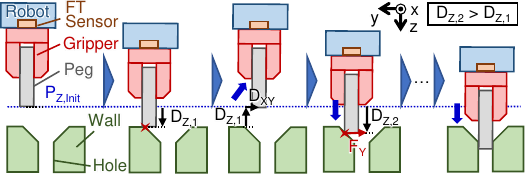}\\[-2ex]
  \subfloat[\label{subfig:concept1}]{\hspace{.2\columnwidth}}
  \subfloat[\label{subfig:concept2}]{\hspace{.2\columnwidth}}
  \subfloat[\label{subfig:concept3}]{\hspace{.2\columnwidth}}
  \subfloat[\label{subfig:concept4}]{\hspace{.2\columnwidth}}
  \subfloat[\label{subfig:concept5}]{\hspace{.2\columnwidth}}
  \caption{Hole search method. $P_{z,init}$ is the initial offset position. $D$ is the peg displacement. Hole borders are chamfered due to concrete's brittleness. (a) Approach. (b) Attempt. (c) Separation. (d) Attempt. (e) Insertion.}
  \label{fig:concept}
  \end{center}
  \end{figure}

The hole search method used is shown in Fig. \ref{fig:concept}. It consists of the following steps: (a) moving the peg to a hole position roughly estimated by a camera; (b) attempting to insert the peg by moving it toward the hole; (c) separating the peg from the wall and moving it to the next search position if the peg touches the wall; (d) attempting to insert the peg again; and repeating steps (c) and (d) until (e) the hole is found. By separating the peg from the wall, the unwanted effect of the high friction coefficient of the concrete surface can be avoided when moving the peg between hole search positions. In anchor-bolt insertion, after the hole is found, the peg (i.e., anchor bolt) is completely inserted into the wall by hammering, which overcomes the peg ``wedging'' and ``jamming'' problems considered in classic studies \cite{peginhole_model1}.

\subsection{Motion generation model}\label{subsec:model}

\subsubsection{Model inputs and outputs}\label{subsubsec:model_input_output}
 
To control the robot to move toward the hole, we propose the model shown in Fig. \ref{fig:saprle2e}. The model, namely the reinforcement learning (RL) agent, is given a set of six sequential observations $o_{t-5 \twodots t}$ of the environment (time t-5 to t). This window of observations is used because: (1) fully connected layers are used to generate sequential robot motion instead of LSTM as in \cite{ichiwara_icra22}, (2) it reduces computational cost as the LSTM's back propagation through time is avoided, (3) it is effective for inferring temporal statistics \cite{drl_atari}, and (4) it simplifies data storage and random batch sampling from the replay buffer to train the agent. Each observation consists of an image $i$, forces and torques $FT$, robot displacement $D_z$, and the previous robot action $a$. $FT$ includes the forces in the x, y, and z axes and the moments in the x and y axes. $D_z$ is used because it increases as the peg approaches the hole since the brittle borders of holes in concrete often become chamfered. Previous actions are used because stochastic policies should also condition their action on the action history \cite{action_importance}, and the previous actions prevent the robot from looping back to the same place \cite{andre_sii2022}. After the model is input with this window of observations, it predicts the images of the next time $\hat{i}_{t-4 \twodots t+1}$ and the actions $a(dir,ss)_{t+1}$ to be taken by the robot (RL environment). The actions are peg movements in one of eight directions ($dir$ = up, down, left, right, or diagonals) and one of three discrete step sizes ($ss$ = 1, 2, or 3 mm), for a total of 24 different action options.

\begin{figure}[t]
  \begin{center}
  \includegraphics[width=\columnwidth]{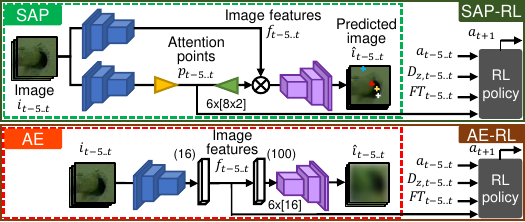}
  \caption{Details of SAP-RL (top) and AE-RL (bottom).}
  \label{fig:saprl}
  \end{center}
\end{figure}

\subsubsection{Model structure}\label{subsubsec:model_struct}
The proposed model consists of a spatial attention point network (SAP) integrated to a deep RL policy (Fig. \ref{fig:saprle2e}). Both SAP and the RL policy are trained end-to-end; thus, we call it SAP-RL-E. Unlike prior work \cite{ichiwara_arxiv}, we refer to SAP as the network that extracts the attention points. Thus, we removed the LSTM layer of the prior model so SAP could work alone (Fig. \ref{fig:saprl}) or with the RL policy (Fig. \ref{fig:saprle2e}). SAP consists of an attention point extraction block (hereinafter, attention block), an image feature extraction block, and an image prediction block. The attention block uses a CNN-based encoder block and soft argmax to extract the positional coordinates of maximum activation from an input image \cite{dsae}, which are the attention points $p$. The image feature extraction block also extracts features $f$ of the input image with an encoder block and inputs it into the image prediction block in a skip connection manner. The attention points from the current time $p_{t-5 \twodots t}$ that are output by the attention block are input to the RL policy together with $FT$, $D_z$, and $a$ (hereinafter referred to as proprioceptive data). The RL policy predicts the attention points of the next time $\hat{p}_{t-4 \twodots t+1}$ as well as the action to be taken by the robot $a_{t+1}$. The image prediction block uses the predicted attention points $\hat{p}_{t-4 \twodots t+1}$ and the image features $f_{t-5 \twodots t}$ to predict (i.e., reconstruct) the image of the next time $\hat{i}_{t-4 \twodots t+1}$. This block does this by: (i) generating a heat map with an inverse argmax modified to be differentiable (heat map generator), (ii) multiplying the heat map element-wise with $f_{t-5 \twodots t}$, and (iii) passing the result image through a transposed CNN block (decoder) that generates $\hat{i}_{t-4 \twodots t+1}$. The images are reconstructed because it was shown that the reconstruction improves the model performance \cite{reconstruct_eval}, stabilizes the attention points \cite{ichiwara_arxiv}, and avoids manual labelling the points \cite{ichiwara_icra22}. This end-to-end approach makes it possible to simultaneously train SAP and the RL policy, enabling SAP to predict attention points that are relevant to the task and the RL policy to predict RL actions that are guided by those attention points. 

\subsubsection{DRL algorithm}\label{subsubsec:algo}
The DRL algorithm used for the RL policy was the double deep q-network (DDQN) as its effectiveness for this task was shown in our prior work \cite{andre_icra2021}. The hyper-parameters used are shown in Table \ref{tab:hyperparameters}. Under a given policy $\pi$, Q-values are estimated by $Q_{\pi}(s,a)= \mathbb{E}_\pi [\sum_{t = 0}^{\infty} \gamma^t r_{t}|s_t=s, a_t=a]$, where $r_t$ is the reward at time $t$, $\gamma \in [0,1]$ is the discount factor, $a$ is the action, and $s$ is the input state of the DDQN ($p$ + proprioceptive data of total size 3x23). The DNN of DDQN is trained to approximate the optimal Q-function $Q^*(s,a)=\max_\pi{Q_{\pi}(s,a)}$, by updating its weights $\theta$ with the Bellman equation given by $\theta_{t+1} = \theta_t + \alpha \nabla J$, where $\nabla$ is the gradient function, $\alpha$ is the learning rate, and $J$ is the loss function which will be presented in Section \ref{subsubsec:loss}. The policy used to choose the actions based on the Q-function was the Boltzmann exploration policy, in which parameter $\tau$ was set to induce exploration at the beginning of the training and annealed every 50 episodes. A replay buffer was used to store images and proprioceptive data to provide random samples that stabilize DNN training. The reward function implemented for each step is $r=-1$ when the DRL episode does not end and is given by the following when the episode ends:

\begin{align} \label{eq:reward}
  r &=
	\begin{cases}
		r_{found hole},& \resizebox{.17\columnwidth}{!}{$\textrm{if hole found,}$}\\
		0,& \resizebox{.4\columnwidth}{!}{$\text{if }  d \leq d_0 \text{ and hole not found,}$ } \\
		-r_{found hole}\cdot \frac{d-d_0}{d_{lim}-d_0} ,& \resizebox{.4\columnwidth}{!}{$\text{if }  d > d_0 \text{ and hole not found,}$ }
	\end{cases}
\end{align}

Here, $r_{foundhole}$ is the reward when the hole is found, $d_0$ is the initial distance from the hole, $d$ is the final distance, and $d_{lim}$ is the distance limit. The negative reward at the end of each step encourages the DNN to minimize the number of steps. The comparison of $d_0$ and $d$ at the end of the episode induces the model to approximate the distance from the peg to the hole position. The DRL episode ends when (i) the peg reaches distance limit $d_{lim}$ from the hole center, (ii) the number of steps taken exceeds 100, or (iii) the peg is inserted into the hole which is identified when $F_z>F_{z,th}$ and $D_z> D_{z,th}$ (see Table \ref{tab:hyperparameters}). Note that our model is not limited to DDQN and can be used with complex algorithms such as DDPG and SAC. However, more time and elaborate training methods for continuous actions are required \cite{sacae}.

\subsubsection{Loss function}\label{subsubsec:loss}
The loss function is defined as:
\begin{align}
  J&= w_i \cdot J_{i} + J_{Q} + w_p \cdot J_{p}, \label{eq:TotalLoss} \\   
  J_{i}&=\frac{1}{H \cdot W\cdot C \cdot L}\|\hat{i'} - i' \|_2^2, \label{eq:ImgLoss} \\
  J_{Q}&=\frac{1}{2}( r +  \gamma Q^{-}(s',\argmax_a Q(s',a)) - Q(s,a))^2, \label{eq:Qloss} \\
  J_{p}&=\frac{1}{K\cdot L}\| p - \hat{p}'\|_2^2. \label{eq:FeatureLoss}
\end{align} 

Here, $i$ is the image, $p$ is the attention points, $Q$ is the q value of the main DDQN, $Q^{-}$ is the q value of the target DDQN (a copy of the main DDQN), and $r$ is the reward of the RL algorithm \cite{doubleqlearning}. The variables with an apostrophe are the variables of the next time. $J_i$ and $J_{Q}$ are the mean square error of the image prediction and the temporal difference (TD), respectively. $J_p$ is the mean square error of the Euclidean distance between the attention points of the RL policy input $p$ and output $\hat{p}'$, and $w_i$ and $w_p$ denote the loss weight of $J_i$ and $J_p$, respectively. By adding $J_p$, the model can learn to predict attention points in positions near their predecessors. This reflects the reality since the position of important points in the images do not change significantly in one time-step. In this study, RGB images with a resolution of $64\times64$ were used (height($H$) = 64, width($W$) = 64, and channels($C$) = 3). There are eight attention points, and because they are xy coordinates, $K=2\times8$. The window size $L$ is six, and $w_i$ and $w_p$ are annealed from 0.0001 to 0.1 and 1.5 to 1.0, respectively, within episodes 0 and 1000. Starting with a high $w_i$ encourages the model to learn to encode images more optimally before learning to generate motion.

\begin{table}[t]
  \begin{center} 
  \caption{Hyperparameters used by DRL algorithm}
  \label{tab:hyperparameters} 
  \setlength\tabcolsep{4pt} % default value: 6pt
  \begin{tabular}{rc|rc} \hline  	  
  \bf{Batch size} & 32  & \bf{Optimizer}			& Adam	\\
  \bf{Activation hidden layers}		& ReLU		& \bf{Learning rate  ($\alpha$)}		& 0.001	\\ 
  \bf{Activation last layer}		& Linear		& \bf{Discount factor ($\gamma$)}	& 0.99	\\ 
  \bf{Boltzmann parameter ($\tau$)} & 100-1	& \bf{Distance limit ($d_{lim}$)}		& 4.5 mm	\\ 
  \multicolumn{3}{r}{\bf{Episodes for updating target network}}  & 100  \\  
  \multicolumn{3}{r}{\bf{Reward when hole is found ($r_{foundhole}$)}}	& 100	\\ 
  \multicolumn{3}{r}{\bf{Force threshold on z-axis ($F_{z,th}$)}}	& 20 N	\\ 
  \multicolumn{3}{r}{\bf{Displacement threshold on z-axis ($D_{z,th}$)}}	& 9 mm	\\ \hline
  \end{tabular}
  \end{center}
\end{table}

\begin{figure}[t]
  \begin{center}
  \includegraphics[width=\columnwidth]{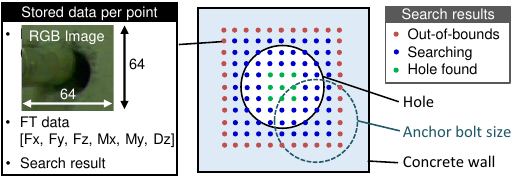}
  \caption{Illustration of hole map for offline training. In this study, area is \textpm 5 mm from hole center and points are spaced 0.25 mm apart.}
  \label{fig:map}
  \end{center}
  \end{figure}

\subsection{Offline training framework}\label{subsec:offtrain}

Because short lead time and robustness are the key requirements for applying the proposed model to construction, we propose a framework that enables offline training of the proposed model with minimal data extraction from the environment. The offline training reduces training time, and the data extraction reduces the reality gap when using the robot in the real world. The framework is made up of three steps: (i) attempting anchor-bolt insertion at multiple points around the hole for a single hole (Fig. \ref{fig:map}); (ii) storing observations (images, proprioceptive data) and search results (``hole found'', ``searching'', or ``out-of-bounds'') from each attempt along with the position of the attempt in order to create a hole map; and (iii) using the hole map data as input to train the model via DRL in place of the real robot.

This hole map can be created because, in this study, the robot moves in discrete steps, and such movement limits the positions around the hole that the robot (i.e., peg) can touch. Thus, the robot movement enables hole mapping that is data efficient. The mapping is performed by attempting insertions following a decided trajectory that is shown in the supplementary video. In contrast to our prior work\cite{andre_sii2022}, the hole map of this study also stores images in addition to the proprioceptive data.

\subsection{Models for comparison}\label{subsec:compmodels}
For comparison with the proposed model, a model that uses only proprioceptive data (P-RL), one that uses an autoencoder (AE) trained separately from the RL policy (AE-RL), and one that uses SAP also trained separately from the RL policy (SAP-RL) were used. P-RL only consists of the RL policy and is trained in conditions similar to that of our prior work \cite{andre_icra2021,andre_sii2022}. AE-RL has its AE trained first with the images of a hole map, and then it uses the trained AE to generate image features to train the RL policy (see Fig. \ref{fig:saprl}). In SAP-RL, SAP is trained in the same manner as that of AE and uses the trained SAP to generate attention points to train the RL policy. The CNNs of SAP and AE are of the same structure as the CNNs of SAP-RL-E. The RL policy of both AE-RL and SAP-RL are also input with proprioceptive data because it improves the network performance during target occlusions as shown in \cite{ichiwara_icra22}. Note that unlike the proposed model, both the AE and SAP learn to generate image features or attention points of the current image and do not consider the robot motion for the predictions. All comparison models are trained with a window size of six similar to the proposed model.

\section{Experimental Setup and Conditions}\label{sec:setupandconditions}
\subsection{Experimental setup}\label{subsec:setup}
\begin{figure}[t]
  \begin{center}
  \includegraphics[width=\columnwidth]{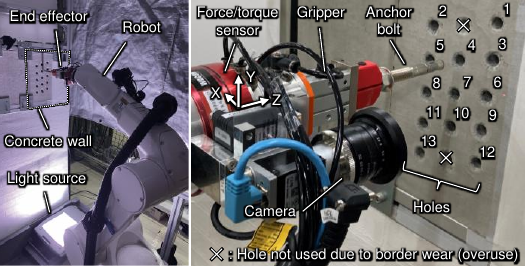}\\[-2ex]
  \subfloat[\label{subfig:denso1}]{\hspace{.4\columnwidth}}
  \subfloat[\label{subfig:denso2}]{\hspace{0.6\columnwidth}}
  \caption{Experimental setup for inserting anchor bolt. (a) Entire setup. (b) End effector, holes, and anchor bolt.}
  \label{fig:setup}
  \end{center}
\end{figure}

The experimental setup used is shown in Fig. \ref{fig:setup}. The setup includes a Denso robot (VM-60B1) equipped with an air gripper (Airtac HFZ20), a force-torque (FT) sensor (DynPick\textsuperscript{\tiny\textregistered} WEF-6A1000-30), and a camera with a wide-angle lens (Basler acA1920-40gc/ Kowa Lens LM6HC F1.8 f6 mm 1"). The camera was set with the default parameters, but its region of interest was set to capture only the area around the peg tip (64x64 px). The robot was programmed to grasp a wedge-type anchor bolt and perform the peg-in-hole task in 13 holes in a concrete wall. The holes were opened with a drill bit of the same diameter as a 12-mm anchor bolt (about 0.2 mm clearance between peg and hole).

During the hole search, the robot was moved following the flow presented in Subsection \ref{subsec:holesearch}. The forces and moments were measured by the FT sensor, and $D_z$ was calculated from the xyz position of the tip of the anchor bolt obtained through forward kinematics. Before each DRL training episode, the robot was placed in a home position perpendicular to the wall, the FT sensor was zeroed (to compensate for gravity), and then the robot was moved to an initial search position. After the episode ended, the robot was retracted and the aforementioned steps were repeated until the maximum number of episodes was reached. 

\begin{figure}[t]
  \begin{center}
  \includegraphics[width=\columnwidth]{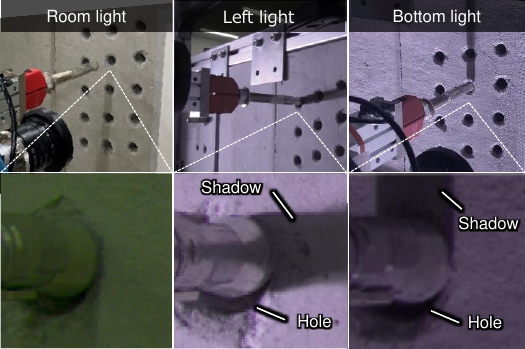}
  \caption{Lighting conditions. Peg casts shadow from left and bottom light.}
  \label{fig:light_cond}
  \end{center}
\end{figure}

The three lighting conditions used are shown in Fig. \ref{fig:light_cond}. They replicate the variations of lighting that occur in the field, which is our definition of variable lighting conditions. The room lighting, generated by multiple sources of white light on the ceiling, replicates a well illuminated field and provided the setup with images with almost no shadow. The left and bottom lighting conditions, generated by a flood light (DENSAN PDS-C01-100FL) set to emit white light at the left and bottom side of the peg respectively, provided images with misleading shadows. These two lighting conditions were chosen because they were considered the most challenging. They are possible if a light source external to the robot system is used. Right and top lights were not used because they did not cast shadows that were visible to the camera.

The PC used for training was equipped with an Intel i7-8700 CPU, 16GB of RAM memory and an NVIDIA GeForce GTX 1070Ti GPU with 8 GB of memory.

\subsection{Training conditions}
To train the models, only the map of hole 1 taken in regular room light was used, which is in accordance with the proposed training framework. They were trained offline for 4000 episodes, starting from initial positions randomly chosen within the ``searching'' search result illustrated in Fig. \ref{fig:map}. To avoid overfitting, Gaussian noise was added to the inputs.

\subsection{Evaluation conditions}

\begin{figure}[tb]
  \begin{center}
  \includegraphics[width=\columnwidth]{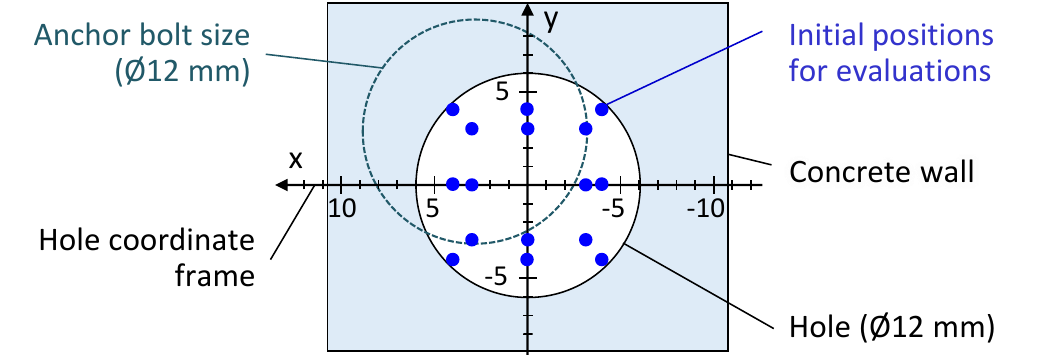}
  \caption{Initial peg positions for method evaluations}
  \label{fig:initpos}
  \end{center}
\end{figure}

To evaluate the proposed model, offline tests (with the hole maps) and online tests (with the real robot) were conducted with all models. Both tests were conducted by attempting to find 12 unknown holes (2 to 13) in all three lighting conditions with the peg starting from the initial positions illustrated in Fig. \ref{fig:initpos}. For the offline tests, hole maps were extracted from all the 12 holes. The tests were repeated ten times per initial position (160 tests per hole per lighting condition, for a total of 5760 tests per model).

The initial positions were set at an absolute distance of 3 and 4 mm away from the hole origin in the x, y, or both axes (hereinafter referred to as initial positions at 3 and 4 mm, respectively). This is because, although the maximum initial positioning error estimated for the current setup is 3 mm, distances that are larger than this error have also to be overcame by the models in field applications.

\section{Results}\label{sec:results}

\subsection{Training results}

\begin{figure}[t]
\begin{center}
\includegraphics[width=\columnwidth]{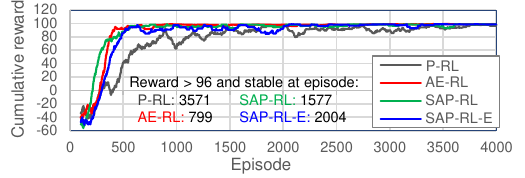}
\caption{Offline training results of RL policy (average of 100 episodes).}
\label{fig:trainres}
\end{center}
\end{figure}

The offline training results are shown in Fig. \ref{fig:trainres} in terms of the average cumulative reward of 100 episodes (hereinafter, cumulative reward). The figure also lists the number of episodes in which the cumulative reward was above 96 (the maximum reached by P-RL) and stable. Overall, all models nearly reached the maximum cumulative reward of 100 during training, demonstrating that the models learned to control the robot to find the holes effectively. However, P-RL took more episodes to converge (3571 episodes), which was expected since it is provided with less input information on the environment. The cumulative reward of AE-RL, SAP-RL, and SAP-RL-E (proposed model) increased quickly, but AE-RL and SAP-RL converged more quickly than SAP-RL-E. This was also expected because, while AE-RL and SAP-RL are trained with image features from pre-trained image encoding networks, SAP-RL-E is trained directly from the image input, a condition which was previously observed to require a longer training time \cite{deepmind}. Nevertheless, considering the high number of episodes for P-RL to converge and the fact that SAP-RL-E trains the image encoding networks in parallel with the RL policy, it can be inferred that SAP-RL-E is sample efficient as its cumulative reward increased quickly and converged with only few episodes more than SAP-RL.

The time required to train the models is shown in Table \ref{tab:traintime}. ``RL train'' is the RL policy training time until convergence (i.e., until a stable cumulative reward above 96 is reached). The online training time, estimated only for P-RL (online P-RL) from the time taken by the robot during online tests, was considerably longer than the time needed to train the other models with the offline training framework. This demonstrates that the framework reduces the training time considerably. The time needed to train SAP-RL-E is almost the same as the time taken by P-RL and AE-RL, which suggests they have similar computational costs. This is because, unlike AE-RL and SAP-RL, SAP-RL-E does not require pre-training an image encoding network, and it takes fewer number episodes to converge than P-RL (2004 versus 3571 for P-RL, as shown in Fig. \ref{fig:trainres}). This fewer number of episodes compensates for the longer time SAP-RL-E takes to train each episode. As a result, the proposed model can easily substitute P-RL at construction sites considering the lead time. Note that the training time of AE is shorter than of SAP because AE has fewer weights (2072 versus 3176 for SAP).

  \begin{table}[t]
    \begin{center} 
    \caption{Training times in hours and minutes. $\diamond$ is the proposed model. * are estimates. Best results is in bold}
    \label{tab:traintime} 
    \setlength\tabcolsep{2.5pt} % default value: 6pt
    \begin{tabular}{rc|c|c|c||c}
      \textbf{Model}              & \textbf{Input}  & \textbf{Mapping}   & \textbf{Pre-train} & \textbf{RL train} & \textbf{Total}  \\ \hline
      \textbf{Online P-RL}        & \textbf{Proprioceptive}    & -                  & -                  & 15:04*            & 15:04*         \\
      \textbf{P-RL}               & \textbf{Proprioceptive}    & 1:10               & -                  & 1:10              & 2:20            \\
      \textbf{AE-RL}              & \textbf{Im+Proprio.} & 1:10               & 0:58               & 0:15              & 2:23            \\
      \textbf{SAP-RL}             & \textbf{Im+Proprio.} & 1:10               & 2:05               & 0:27              & 3:42            \\
      $\diamond$\textbf{ SAP-RL-E}& \textbf{Im+Proprio.} & 1:10               & -                  & 1:09              & \textbf{2:19}   \\
      \end{tabular}
    \end{center}
    \end{table}

\subsection{Offline evaluation results}

The offline test results are listed in Table \ref{tab:offres}. In the room light, all vision-based models outperformed the model that only used proprioceptive feedback (P-RL). They presented higher success rates (SRs) and shorter completion times (CTs). In particular, AE-RL presented a considerably short CT of less than three steps (2.5 s/step), which is quite close to the performance of humans. However, AE-RL presented low SR in harsh lighting conditions, making it unsuitable for construction sites. The models that used SAP (SAP-RL and SAP-RL-E), however, demonstrated robustness to the challenging lighting conditions; namely, they could execute the peg-in-hole task with high SR and short CT, outperforming P-RL in both metrics. The proposed model (SAP-RL-E) yielded the most optimal results with an average SR of 97.4\% and CT of 7.65 s, which is a 9.2\% higher SR and 4.43 s shorter CT than that of the baseline P-RL. These results suggest that the SAP enables the extraction of relevant points of the image even under harsh lighting conditions, and the proposed SAP-RL-E can substitute the P-RL for executing peg-in-hole tasks in the construction field.

\begin{table}[t]
  \begin{center} 
  \caption{Offline test results. $\diamond$: Proposed model,  CI: 95\% confidence interval, $\mu$: mean, E: error. Best results are in bold}
  \label{tab:offres} 
  \setlength\tabcolsep{2.0pt} % default value: 6pt
  \begin{tabular}{r|cc|cc|cc||cc}
  \textbf{Light}& \multicolumn{2}{c|}{\textbf{Room}}       & \multicolumn{2}{c|}{\textbf{Left}} & \multicolumn{2}{c||}{\textbf{Bottom}} & \multicolumn{2}{c}{\textbf{All}}     \\ \hline
  \multirow{2}{*}{\backslashbox[5.5em]{\textbf{Model}}{\textbf{Metric}}}  & \textbf{SR} & \textbf{CT} & \textbf{SR} & \textbf{CT} & \textbf{SR} & \textbf{CT} & \textbf{SR[\%]} & \textbf{CT[s]}  \\ 
  & \textbf{[\%]} & \textbf{[s]} & \textbf{[\%]} & \textbf{[s]} & \textbf{[\%]} & \textbf{[s]}  & \textbf{CI(\boldmath${\mu}$\textpm E)}  & \textbf{CI(\boldmath${\mu}$\textpm E)}  \\ \hline
  \textbf{P-RL}       & 88.2                & 12.08           & -             & -             & -               & -              & 88.2 \textpm   1.4         & 12.08         \textpm 0.46  \\
  \textbf{AE-RL}      & 91.4                & \textbf{6.30}   & 22.6          & \textbf{3.17} & 59.5            & 8.42           & 57.8 \textpm   1.3         & \textbf{5.96} \textpm 0.16  \\
  \textbf{SAP-RL}     & 95.1                & 9.57            & 93.6          & 10.25         & 95.3            & 9.23           & 94.7 \textpm   0.6         & 9.68          \textpm 0.21  \\
  $\diamond$\textbf{SAP-RL-E}& \textbf{97.4}& 7.37            & \textbf{95.8} & 7.76          & \textbf{99.1}   & \textbf{7.82}  & \textbf{97.4} \textpm 0.4  & 7.65          \textpm 0.13  \\
  \end{tabular}
  \end{center} 
\end{table}

\subsection{Online evaluation results}
Table \ref{tab:onres} lists the online test results of the models. The tests with the real robot showed that, as in the offline test results, the proposed SAP-RL-E model was the most successful overall. AE-RL had the lowest CT, but its low SR in challenging lighting makes it ineffective in the field. The success rate of P-RL decreased the farther the peg initial position was from the hole center (average success rate of the initial positions 4 mm from the hole was 79.4\%). We hypothesize that this is because the forces and moments of the x and y axes (in the wall plane) become smaller the farther the peg is from the hole center. This issue did not greatly affect the vision-based models, which suggests they can generalize more effectively to different initial positions.

Although SAP-RL-E and SAP-RL performed similarly, their confidence interval (CI) did not overlap. Also, the p-values for SR and CT calculated via two-tailed z-tests were 0.026 and 0.002 ($<$0.05), respectively. These results suggest that the differences between the results of the models are statistically significant; thus, SAP-RL-E is more effective than SAP-RL.

\begin{table}[t]
  \begin{center} 
  \caption{Online test results. $\diamond$: Proposed model,  CI: 95\% confidence interval, $\mu$: mean, E: error. Best results are in bold}
  \label{tab:onres} 
  \setlength\tabcolsep{2.0pt} % default value: 6pt
  \begin{tabular}{r|cc|cc|cc||cc}
  \textbf{Light}& \multicolumn{2}{c|}{\textbf{Room}}       & \multicolumn{2}{c|}{\textbf{Left}} & \multicolumn{2}{c||}{\textbf{Bottom}} & \multicolumn{2}{c}{\textbf{Average}}     \\ \hline
  \multirow{2}{*}{\backslashbox[5.5em]{\textbf{Model}}{\textbf{Metric}}}  & \textbf{SR} & \textbf{CT} & \textbf{SR} & \textbf{CT} & \textbf{SR} & \textbf{CT} & \textbf{SR[\%]} & \textbf{CT[s]}  \\ 
  & \textbf{[\%]} & \textbf{[s]} & \textbf{[\%]} & \textbf{[s]} & \textbf{[\%]} & \textbf{[s]}  & \textbf{CI(\boldmath${\mu}$\textpm E)}  & \textbf{CI(\boldmath${\mu}$\textpm E)}  \\ \hline
  \textbf{P-RL}                   & 87.4          & 11.57          & -             & -             & -             & -             & 87.4          \textpm 1.5  & 11.57         \textpm 0.35  \\
  \textbf{AE-RL}                  & 91.4          & \textbf{6.40}  & 19.4          & \textbf{3.88} & 57.1          & 8.34          & 56.0          \textpm 1.5  & \textbf{6.21} \textpm 0.19  \\ 
  \textbf{SAP-RL}                 & 92.8          & 8.78           & \textbf{92.6} & 10.26         & 92.2          & 9.01          & 92.5          \textpm 0.7  & 9.35          \textpm 0.20  \\ 
  $\diamond $\textbf{SAP-RL-E}    & \textbf{94.5} & 7.85           & 91.6          & 8.62          & \textbf{95.5} & \textbf{8.15} & \textbf{93.9} \textpm 0.7  & 8.21          \textpm 0.22
  \end{tabular}
\end{center} 
\end{table}

\subsection{Image reconstruction and attention point analysis}
\begin{figure}[t]
  \begin{center}
  \includegraphics[width=\columnwidth]{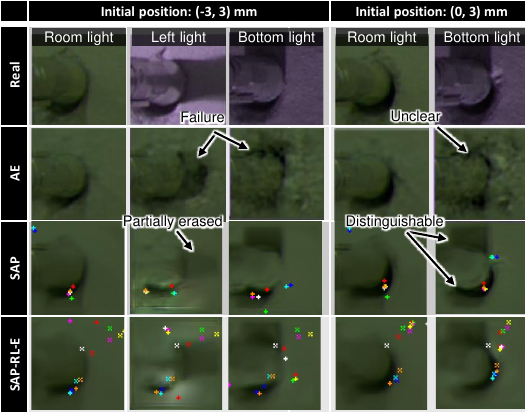}
  \caption{Image reconstruction and attention point results. SAP generates only eight attention points related to the current image (cross-shaped marks), while SAP-RL-E generates additional eight attention points related to the predicted images (x-shaped marks).}
  \label{fig:img_recons}
  \end{center}
\end{figure}

Fig. \ref{fig:img_recons} presents the image reconstruction and attention points generated by the vision-guided models. In the room light, all models were able to reconstruct the images adequately, and SAP and SAP-RL-E were able to generate attention points focused on the region between the peg and the hole (particularly the blue and orange points for SAP-RL-E). However, in the other lighting conditions, AE failed to reconstruct the image adequately, either predicting the hole to be in the shadow position or failing to reconstruct a clear image. In contrast, SAP-RL and SAP-RL-E were able to reconstruct the image clearly; they partially ``erased'' the shadow or created an image in which the shadow and the hole positions were clearly distinguishable. The positions of attention points of SAP-RL-E were also minimally affected by the shadows. The attention points from the current (cross-shaped) and predicted images (x-shaped) of SAP-RL-E provided a sense of direction for the robot's movement. The combination of these attention point-based directions may be related to the robot’s movement direction, but this will need to be verified in future work. The results suggest that SAP was most effective for reconstructing the image and generating attention points relevant to the task.

\section{Discussion}

\subsection{Application to construction}

Although it was demonstrated that the proposed model is the most suitable for a wide range of construction environments, different models can be chosen depending on the environment. If a light source is placed close to the camera to reduce the influence of challenging external lights (e.g., dawn light), the AE-RL may be chosen because of its shorter CT. However, preliminary evaluations of AE-RL with a camera light showed that the SR dropped more than 20\% when an external light was added, so the proposed model still yields a higher SR. 

If the lighting conditions are extremely poor, P-RL can be chosen since it is independent of lighting conditions, given that the robot can make contact with the hole borders without vision (e.g., by using CAD-obtained hole positions). However, it is difficult to accurately estimate position without vision because construction environments are highly unstructured. Thus, if vision is already used to approach the hole, the choice of a vision-based hole search model is more optimal.

\subsection{Hole map size}

The hole map size affects the mapping time and the image encoding capability of the image encoding networks (AE and SAP). If the map is large,  due to a higher map resolution, for example, more time is required for mapping (extracting data). However, in this case, more images are available for training the image encoding networks, enabling them to learn to encode the images more robustly \cite{deep_learning}. On the contrary, if the hole map is small, the mapping time is short, but the image encoding networks are not able to adequately encode the input images. Small resolution maps with interpolation functions could enable more detailed maps with less data extraction, but this approach creates synthetic data that may be different from reality, resulting in reduced robot performance in the real world. The results of this study suggest that the map size used provided a balance between mapping time and image encoding capability, which led to robust peg-in-hole task execution. To further improve the robustness of the proposed model, multiple holes maps can be used to train a model.

\section{Conclusion}

We proposed a vision and proprioceptive data-driven model to accomplish the peg-in-hole task in concrete holes under challenging lighting conditions. The model uses a spatial attention point network (SAP) and a RL policy trained jointly end-to-end in order to extract task-specific attention points from images and generate robot motion based on those attention points and proprioceptive data. We also proposed a training framework that involves mapping a target hole and using this hole map to train the model via DRL in an offline manner. The hole mapping minimizes the reality gap when transferring the model to the real world, and the offline training reduces training time. Through evaluations with an experimental setup that replicates a construction environment, we demonstrated that by training the proposed model with a single hole map and in room light, the model can successfully control an industrial robot to perform the peg-in-hole task in 12 unknown holes, starting the peg from 16 different initial positions and under three different lighting conditions (two with misleading shadows). The proposed model reached a success rate (SR) of 93.9\% and completion time (CT) of 8.21 s, outperforming a series of baseline models, including one that only used proprioceptive data as input (SR:87.4\%, CT:11.57 s). Although the proposed model was used for anchor-bolt insertion, it may be applied to similar tasks where peg sliding problems due to complex surfaces and light variations can occur, such as insertion of pins in furniture, or connectors in circuits in poorly lit assembly lines.

\bibliographystyle{IEEEtran}
\bibliography{myBib}

\end{document}